\documentclass[sigconf, nonacm=true]{acmart}
\usepackage{multirow}\usepackage{subcaption}
\usepackage{algorithm}
\usepackage{algpseudocode}
\usepackage{enumitem}
\usepackage{longtable}
\usepackage{tcolorbox}
\AtBeginDocument{%
  }

\setcopyright{acmlicensed}
\copyrightyear{2018}
\acmYear{2018}
\acmDOI{XXXXXXX.XXXXXXX}
\acmConference[Conference acronym 'XX]{Make sure to enter the correct
  conference title from your rights confirmation email}{June 03--05,
  2018}{Woodstock, NY}
\acmISBN{978-1-4503-XXXX-X/2018/06}

\begin{document}

\title{FineServe: A Fine-Grained Dataset and Characterization of Global LLM Serving Workloads}

\author{Tiancheng Zhang}
\authornote{Both authors contributed equally to this research.}
\email{tianchengzhang@tju.edu.cn}
\orcid{0009-0006-7681-1223}
\author{Shaoyuan Huang}
\orcid{0000-0002-4091-6457}
\authornotemark[1]
\email{hsy_23@tju.edu.cn}
\affiliation{%
  \institution{Tianjin University}
  \state{Tianjin}
  \country{China}}

\author{Mingyuan Wang}
\orcid{0009-0002-0508-7451}
\affiliation{%
  \institution{Tianjin University}
  \state{Tianjin}
  \country{China}}
\email{mingyuanwang@tju.edu.cn}

\author{Yunfeng Zhao}
\authornote{Corresponding author.}
\orcid{0000-0001-6767-7233}
\affiliation{%
  \institution{Tianjin University}
  \state{Tianjin}
  \country{China}}
\email{yfzhao97@tju.edu.cn}

\author{Xiaofei Wang}
\orcid{0000-0002-7223-1030}
\affiliation{%
  \institution{Tianjin University}
  \state{Tianjin}
  \country{China}}
\email{xiaofeiwang@tju.edu.cn}

\author{Wenyu Wang}
\affiliation{%
  \institution{PPIO Cloud (Shanghai) Co., Ltd.}
  \state{Shanghai}
  \country{China}}
\email{wayne@pplabs.org}

\renewcommand{\shortauthors}{Zhang and Huang, et al.}

\begin{abstract}
Large language models (LLMs) are increasingly deployed as always-on online services, making efficient LLM serving a critical systems challenge. Achieving low latency and high throughput under volatile demand requires deep understanding of real-world serving workloads, yet existing studies often rely on proxy traces or coarse-grained characterizations that fail to capture the heterogeneity of modern multi-model LLM platforms. 
We present FineServe, an in-the-wild, multi-model LLM serving workload dataset collected from a global commercial marketplace, enabling fine-grained characterization of real-world serving dynamics across heterogeneous models and tasks. Leveraging FineServe, we conduct a comprehensive analysis of arrival dynamics and token behavior, revealing fundamentally different fluctuation regimes across model architectures, scales and task intents.
Building on these insights, we develop the FineServe workload generator, which composes fine-grained model-aware workloads into configurable mixtures tailored for benchmarking multi-model serving platforms.
By exposing these fine-grained workload dynamics, FineServe provides a realistic foundation for evaluating routing, scheduling, and capacity-planning strategies in LLM serving systems. 
FineServe is available at \url{https://github.com/hihiztc1/FineServe}.

\end{abstract}




\maketitle

\section{Introduction}


Large language models (LLMs) have quickly become a central pillar of modern AI services~\cite{deepseekv3, qwen}, powering applications such as code assistants, conversational robots, E-commerce~\cite{ecommerce}, and emerging agentic workflows~\cite{autogen}. 
With the growing deployment of LLMs for always-on online services~\cite{gpt4o}, LLM serving efficiency has turned into a critical systems problem: providers must sustain low latency and high throughput under volatile demand while keeping GPU costs under control.
This has sparked intense efforts to optimize LLM serving stacks~\cite{flashAttention2,flashAttention3}, including scheduling and batching policies~\cite{orca}, KV-cache management, and prefill-decode disaggregation~\cite{distServe}, as well as elastic scaling in multi-model cloud deployments, where performance improvements translate directly into substantial cost savings and better quality of service (QoS).

Realistic serving workload traces are central to this innovation cycle: they not only drive system design by revealing the actual operating regimes, but also provide the credible basis for evaluating routing, scheduling, and capacity-planning strategies.
Existing LLM serving research has relied on adapted or proxy workloads derived from general deep-learning and cloud-computing traces~\cite{HuSC21, JeonATC19, WengNSDI22}, such as treating serverless function invocations as inference requests, as well as more recent approaches combining open-source LLM datasets (e.g., ShareGPT) with synthetic arrival processes~\cite{llumnix}.
While useful, these proxies often impose simplified arrival assumptions (e.g., stationary Poisson-like processes), which miss the non-stationary and bursty behaviors seen in production and can distort system evaluation and derived insights.
Recent efforts have taken important steps toward grounding LLM serving research in real-world traces. BurstGPT~\cite{burst} releases a real-world serving trace and highlights pronounced burstiness and non-stationarity.
ServeGen~\cite{servegen} provides a broader, production-scale characterization across LLMs, emerging multimodal and reasoning models, highlighting that both arrival dynamics and token-length distributions can vary substantially across serving settings.


However, existing traces and characterizations still fall short for today's evolving LLM serving ecosystem in two key aspects:

\textit{(1) Insufficient workload granularity.}
With the rapid proliferation of LLMs, especially high-quality open-source models~\cite{touvron2023llama,bai2023qwen,kimitteam2025kimik2} and increasingly diverse user demands, LLM serving platforms are shifting toward multi-model deployments~\cite{chen2023frugalgpt,ong2025routellm}.
They host heterogeneous model portfolios that differ in architecture (e.g., Dense vs.\ Mixture-of-Experts), scale, capability, and cost.
Meanwhile, user requests span diverse task intents (e.g., writing, coding, reasoning, and role-playing), whose input/output token patterns and latency sensitivities can vary significantly.
Yet many existing traces and analyses~\cite{burst,servegen,stojkovic2025dynamollm,hu2024llmdatacenter} do not provide sufficient visibility into how workload signatures differ across these fine-grained characteristics, making it difficult to reason about multi-model system behaviors.

\textit{(2) Limited capture of market-driven dynamics.}
Most available datasets are collected under provider-specific offerings, where the served model portfolio and user behavior appear relatively stable within curated production settings. In contrast, in market-driven environments, user choice and cost-performance tradeoffs, together with rapid model turnover, can trigger sharp, time-localized shifts in demand. For example, the release of a new model or a sudden change in availability may quickly re-route traffic, causing request surges, shifting token-length profiles, and amplifying system overhead.



As a result, the community still lacks an in-the-wild dataset that enables comprehensive, fine-grained characterization of multi-model and multi-task serving workloads.
We present the first \emph{in-the-wild} multi-model LLM serving workload dataset with \emph{fine-grained system observability} from a global commercial marketplace. Our dataset records privacy-preserving service-level metadata, including timestamps, model identifiers, and input/output token counts enabling systematic characterization along multiple heterogeneity axes: model architecture, scale tier, and task intent.
Using this dataset, we conduct a comprehensive measurement study that organizes the analysis around system-critical dimensions (request arrivals, token geometry, and latency) and distills \textbf{seven} key findings that reveal distinct, axis-dependent workload signatures obscured by aggregate views.
Finally, we introduce \textbf{FineServe}, a lightweight workload generation toolkit that parameterizes fine-grained workloads across models and tasks and composes them into configurable mixtures, supporting both faithful trace replay and scalable synthesis for benchmarking.



In summary, this paper makes the following contributions:
\begin{itemize}[leftmargin=*,itemsep=2pt,topsep=2pt]
    \item We release an in-the-wild, privacy-preserving multi-model LLM serving dataset with fine-grained service observability.
    \item  We provide a fine-grained characterization across model architectures, scale tiers, and task intents, and reveal seven key findings that expose workload signatures in request arrivals and token geometry, directly informing routing, scheduling, and capacity planning for multi-model serving.
    \item We release FineServe, a practical workload generation framework supporting both trace replay and parametric synthesis, that operationalizes these signatures as configurable knobs for realistic, reproducible multi-model benchmarking.
\end{itemize}

\section{Preliminary and Motivation}\label{sec:prelim_motivation}

\subsection{LLM Characteristics in the Wild}
\textbf{Model architecture diversity.}
Modern LLM serving platforms increasingly host models with heterogeneous architectures, primarily Dense Transformers~\cite{transformer, GPT-3} and Mixture-of-Experts (MoE) models~\cite{moe, Transformermoe, deepseekmoe}. Dense models activate all parameters for each token, whereas MoE routes tokens to a subset of experts, leading to distinct system behaviors across serving stages. In the compute-intensive prefill stage, MoE can exploit expert parallelism to achieve higher utilization, while in the memory-bandwidth-bound decoding stage, expert imbalance and cross-device communication may introduce additional contention and latency. Consequently, architectural differences can materially affect throughput, latency and tail-latency sensitivity under bursty, multi-tenant workloads.

\textbf{Model scale diversity.}
Serving workloads also span a wide range of model scales, commonly indexed by parameter count (e.g., 7B/13B/70B/>\!100B) and, in practice, by the resulting hardware footprint (e.g., GPU memory demand, tensor/pipeline parallel degrees, and achievable batch sizes).
Larger models generally increase compute per token and memory pressure, amplify KV-cache footprint during decoding, and tighten batching and scheduling constraints.
In contrast, smaller models can sustain higher concurrency with larger batches, yet their workloads tend to be more demand-driven and volatile, exhibiting sharper and more frequent spikes.

\textbf{Task-intent diversity.}
User requests span diverse task intents such as writing, coding, and role-playing, each inducing distinct prompt structures, input/output token geometry, and latency sensitivity.
For example, interactive chat and role-playing often involve multi-turn exchanges and stricter perceived time-to-first-token requirements, while coding requests may exhibit templated prompts, structured outputs, and longer context carry-over.
These intent-driven differences imply that characterizing serving workloads purely by aggregate arrival rates is insufficient.

\begin{table}[t]
\centering
\caption{Comparison between \textsc{FineServe} and prior characterizations of LLM serving workloads.}
\label{tab:compare_fineserve}
\small
\setlength{\tabcolsep}{4pt}
\resizebox{\columnwidth}{!}{
\begin{tabular}{llll}
\toprule
 & \textbf{FineServe (Ours)} & \textbf{ServeGen}~\cite{servegen} & \textbf{BurstGPT}~\cite{burst} \\
\midrule
\multicolumn{4}{c}{\textbf{Characterization $>$ Scale}} \\
\midrule
Duration          & 4 months &  4 months & 4 months \\
\#Models           & 55       & 12 & 2 \\
\#Series     & 10       & 3 & 2  \\
\#Requests         & 1.48B    & 3.54B & 5.29M \\
\midrule
\multicolumn{4}{c}{\textbf{Characterization $>$ Scope}} \\
\midrule
Architecture               & \checkmark (Dense \& MoE) & $\times$ & $\times$ \\
Scale tiers               & \checkmark (4 bins) & \checkmark (3 bins) & $\times$ \\
Task intents                & \checkmark (10) & $\times$ & $\times$ \\
\midrule
\multicolumn{4}{c}{\textbf{Workload Generation}} \\
\midrule
\multirow[t]{2}{*}{Approach} & Fine-grained           & Parameterized & Parameterized \\
                          & parameterized blending & clients       & burstiness   \\
\bottomrule
\end{tabular}}
\end{table}

\subsection{Towards Fine-grained Serving Workload}


\textbf{Workload characterization and generation.} Optimization of LLM serving systems can yield substantial performance improvements and cost reductions~\cite{sglang,deepspeed_mii,ttrt}, yet doing so hinges on a deep understanding of real-world serving workloads. 
Such understanding is particularly critical for today's multi-model platforms, which simultaneously host heterogeneous model portfolios and diverse task intents.
Table~\ref{tab:compare_fineserve} summarizes representative production-scale characterizations.
While recent datasets provide valuable evidence on request burstiness and token-length behaviors~\cite{burst, servegen}, they still fall short of capturing the fine-grained heterogeneity inherent to multi-model serving systems.
As shown by the comparison, state-of-the-art characterizations do not expose workload signatures jointly along key axes such as \emph{architecture}, \emph{scale tiers}, and \emph{task intents}, and therefore leave many cross-cutting patterns uncharacterized.
Motivated by this, we perform a comprehensive production-scale characterization of in-the-wild serving workloads on a multi-model deployment platform.
Further, we build a workload generation module that parameterizes and fits the sub-workloads induced by fine-grained axis including: architecture, scale, and task-specific components and then blends them into configurable mixtures, enabling practitioners to instantiate realistic workloads that preserve the fine-grained signatures observed in production.


\textbf{Workload source.} 
Our characterization is grounded in real inference workloads collected from a large-scale, global multi-model serving platform built on an open, distributed compute marketplace. The platform maintains a catalog of more than 60 foundation models and continuously incorporates newly released models, leading to rapid demand shifts driven by model quality, pricing, and emerging trends. It aggregates heterogeneous GPU resources across over 4,000 distributed compute nodes worldwide, comprising on the order of $10^4$ GPUs, and exposes unified pay-as-you-go APIs and on-demand instances, including containerized and serverless GPUs, allowing users to invoke diverse models and capacities without managing the underlying infrastructure.

\begin{figure}[htbp]
    \centering
    \includegraphics[width=\columnwidth]{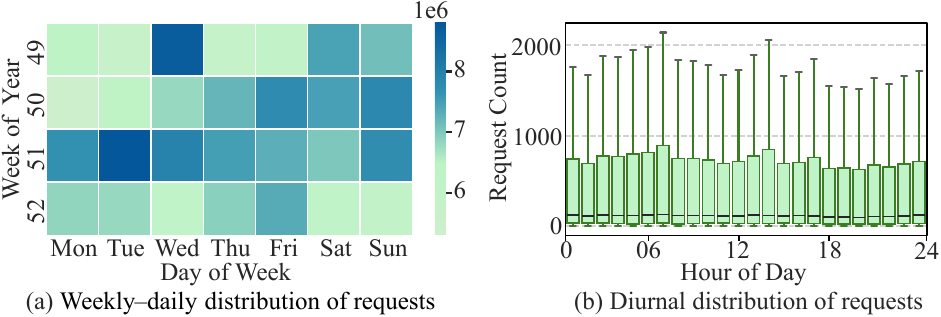}
    \caption{Temporal distribution of FineServe.}
    \label{fig:basic}
\end{figure}

As summarized in Table~\ref{tab:compare_fineserve}, we analyze four months of production traces comprising 1.48B requests across 57 distinct models from 10 model families (detailed in the Appendix~\ref{ap:models_tasks}). 
The traces are collected from the platform’s backend inference logging system and record privacy-preserving service-level metadata for each request, including arrival time, input and output token lengths, and the associated model architecture and scale. 
We analyze Dense models (\texttt{<10B, 10–30B, 30–70B}) and MoE models (\texttt{>100B}), excluding smaller MoE models due to limited samples. All data are fully sanitized and contain no user-identifying content, retaining only metadata-level signals required for workload characterization in compliance with privacy requirements.
As a preliminary overview, Fig.~\ref{fig:basic} presents the weekly request distribution for a representative month and the aggregated daily distribution across all models. Owing to the global coverage of the platform, we do not observe pronounced diurnal or tidal patterns in request volume.

\section{Characterizing Architectures and Scales}
This section characterizes serving workloads across the model \emph{architectures} and \emph{scale tiers} summarized in Table~\ref{tab:compare_fineserve}.
We focus on two critical dimensions: request arrival patterns (\S\ref{sec:arch_scale_arrival}) and input--output token geometry(\S\ref{sec:arch_scale_token}), as they jointly determine queuing pressure, batching opportunities, KV-cache footprint, and ultimately end-to-end performance in LLM serving.

Architectures and scales shape these dimensions through both market and system mechanisms.
On the market side, architecture and parameter size affect inference cost and achievable latency, thereby influencing pricing and user model selection; consequently, different model families experience distinct request arrival patterns rather than a single homogeneous arrival process.
On the system side, model architecture and scale also shape token geometry, as different model families and sizes exhibit distinct input-output length profiles (in particular, output-length variability and tail behavior).

\begin{figure}[tbp]
    \centering
    \includegraphics[width=0.9\columnwidth]{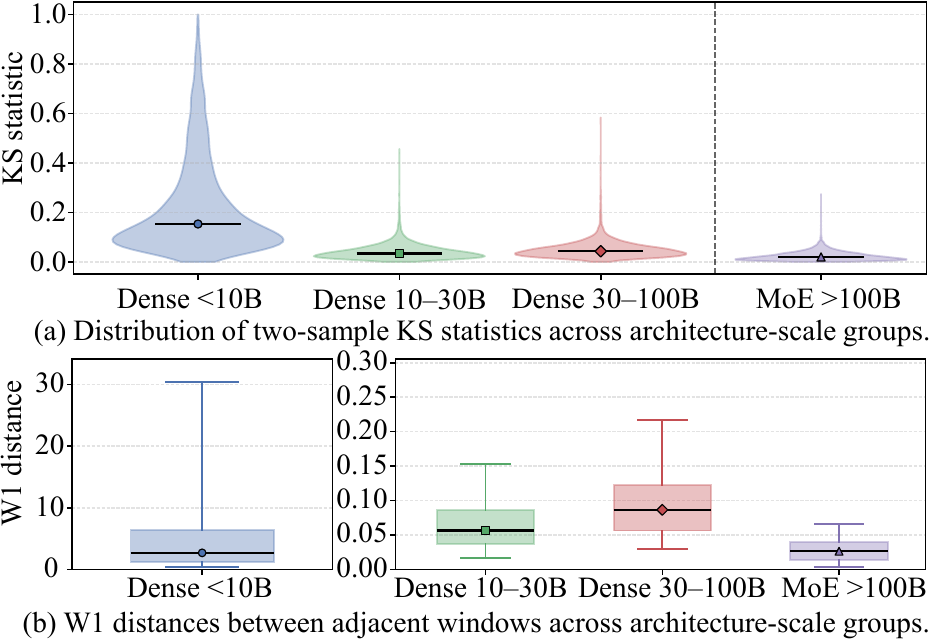}
    \caption{Long-term request distribution shifts measured by two-sample KS statistic and W1 distances.}
    \label{fig:long-term distribution}
\end{figure}

\subsection{Request Arrival Pattern}\label{sec:arch_scale_arrival}
We characterize request arrivals from two complementary perspectives.
1) \emph{long-term distribution shifts} capture how the arrival distribution evolves over days to months, reflecting persistent changes in demand composition that can invalidate stationary workload assumptions and complicate capacity planning. 2) \emph{short-term burstiness and fluctuation} describes within-window and second-level variability, which directly drives real-time scheduling such as queue buildup, batching dynamics, and tail-latency amplification.


\subsubsection{Long-term Distribution Shifts} 
To quantify long-term distribution shifts, we track distributional changes between consecutive time windows over the full observation period.

\textbf{Metrics and Measurement Methodology.} 
Specifically, for each architecture-scale group, we partition the per-second request-count trace into fixed-length windows of $T=300$ seconds (5 minutes).
Let $\mathbf{x}_{i} \in \mathbb{R}^{T}$ denote the request-count samples in $i$-th window:
\begin{equation}
\mathbf{x}_{i} = \{x_{i}(t)\}_{t=1}^{T}, \quad i=0,1,\dots,N-1,
\end{equation}
where $N$ is the total number of windows across the entire measurement period.
We then compare consecutive windows $(i,i{+}1)$ and compute the following metrics:

\emph{(1) Two-sample Kolmogorov--Smirnov (KS) statistic.} Let $\widehat{F}_{i}(\cdot)$ and $\widehat{F}_{i+1}(\cdot)$ denote the empirical cumulative distribution functions (CDFs) of samples in $\mathbf{x}_{i}$ and $\mathbf{x}_{i+1}$, respectively.
We compute the two-sample Kolmogorov--Smirnov statistic as
\begin{equation}
\mathrm{KS}_{i} \triangleq \sup_x \left|\widehat{F}_{i}(x) - \widehat{F}_{i+1}(x)\right|.
\end{equation}
In practice, we use the standard two-sample KS test and record its statistic value $\mathrm{KS}_{i}\in[0,1]$.
A larger $\mathrm{KS}_{i}$ indicates a stronger \emph{distributional discontinuity} between two adjacent windows.

\emph{(2) Wasserstein-1 (W1) distance.}
To quantify the overall magnitude of workload changes, we additionally compute the W1 distance between adjacent windows:
\begin{equation}
\mathrm{W1}_{i} \triangleq W_1\!\left(\widehat{F}_{i}, \widehat{F}_{i+1}\right).
\end{equation}
The W1 distance measures the minimal transport cost to transform the request-count distribution in window $i$ into that in window $i{+}1$.

\textbf{Analysis across Architectures and Scales.} As illustrated in Fig.~\ref{fig:long-term distribution}, Dense models below 10B parameters exhibit the strongest long-term distribution shifts across the two metrics. Their KS statistics are large and heavy-tailed, indicating substantial deviations in distributional shape. 
Besides, the W1 distances reveal that the drift is often severe, with shift magnitudes spanning orders of magnitude.
This behavior aligns with the market positioning of small Dense models. Their low deployment cost and short inference latency make them attractive for a wide range of heterogeneous applications, resulting in broad and disruptive workload shifts.
As model scale increases, Dense models exhibit improved workload stability. Both KS statistics and W1 distances decrease, indicating that distributional changes become less extreme.

MoE models demonstrate a markedly different profile. Their KS statistics and W1 distances remain consistently low, suggesting limited long-term drift in both shape and magnitude. 
These mild distributional adjustments are consistent with their more specialized and higher-cost application scenarios.

\begin{tcolorbox}[
colback=gray!10,
colframe=white,
boxrule=0pt,
arc=0pt,
left=3pt,
right=3pt,
top=3pt,
bottom=3pt
]
\textbf{Finding 1:} Although request arrivals are not directly determined by model architecture, architectural and scale choices shape market adoption and usage patterns, leading to systematic differences in long-term distribution shifts, with Dense models, especially small ones, exhibit larger and more severe distribution shifts, while large MoE models remain comparatively stable.
\end{tcolorbox}



\begin{figure}[htbp]
    \centering
    \includegraphics[width=\columnwidth]{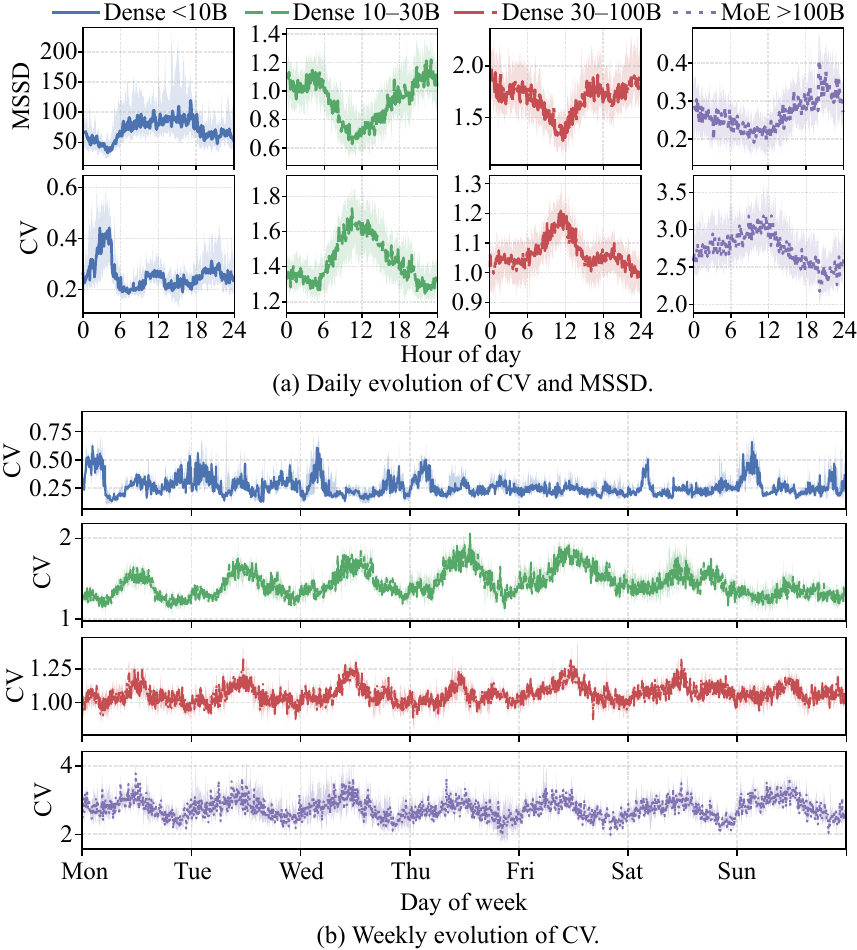}
    \caption{Short-term workload burstiness and fluctuation.}
    \label{fig:burstiness}
\end{figure}

\subsubsection{Short-term Burstiness and Fluctuation} \label{sec:variability}


Long-term drift describes how arrival distributions evolve across time, but serving systems must also react to second-level burstiness and jitter that directly drive queue buildup and tail latency.
In practice, short-term burstiness is multi-faceted: arrivals can be uneven within a window (high peak-to-mean disparity) and highly oscillatory at fine temporal granularity.
We therefore characterize short-term dynamics at per-second resolution using complementary metrics.

\textbf{Metrics and Measurement Methodology.} We quantify short-term burstiness over 5-minute windows ($T{=}300$) using:

\emph{Coefficient of Variation (CV).}
For each window, we compute the CV of per-second request counts, which measures relative dispersion and captures how uneven arrivals are within a window. Notably, CV is order-invariant and does not encode temporal ordering.

\emph{Mean Squared Successive Difference (MSSD).}
To characterize fine-grained fluctuations, we compute
\begin{equation}
\mathrm{MSSD} = \frac{1}{T-1} \sum_{t=1}^{T-1} (x_{t+1} - x_t)^2,
\end{equation}
where $\{x_t\}_{t=1}^{T}$ are per-second request counts in the window.
MSSD measures the step-to-step change in arrivals, i.e., how strongly the request rate fluctuates between adjacent seconds. MSSD is order-aware and sensitive to rapid fluctuations and short-lived spikes.

\textbf{Analysis across Architectures and Scales.}
Fig.~\ref{fig:burstiness}(a) presents diurnal profiles of CV and MSSD
for different architecture-scale groups. Across all groups, both metrics exhibit clear diurnal patterns aligned with user activity cycles. However, the magnitude and nature of burstiness differ substantially.

Dense models below 10B parameters consistently exhibit
\emph{low CV but high MSSD}, indicating that their arrivals are relatively even in magnitude within each 5-minute window, yet fluctuate sharply on a finer timescale (second-level). 
In contrast, MoE models show \emph{high CV but comparatively lower MSSD},
reflecting highly uneven arrivals dominated by large-amplitude bursts, while the second-level fluctuations are smoother.
Interestingly, MSSD reaches its minimum around midday and peaks during late-night hours across all groups. This indicates that high-load periods tend to be temporally smoother, whereas low-load periods are dominated by sporadic and irregular arrivals, which in turn amplify jitter.

Fig.~\ref{fig:burstiness}(b) further shows that these short-term regimes persist over the week.
The ordering of CV across model groups remains remarkably stable across the week: \texttt{Dense $<$10B} maintains the lowest CV, whereas MoE remains the highest.
This stability indicates that the observed burstiness patterns are not driven by isolated anomalous days, but represent inherent properties of how different model classes are used in production.

\begin{figure}[htbp]
    \centering
    \includegraphics[width=\columnwidth]{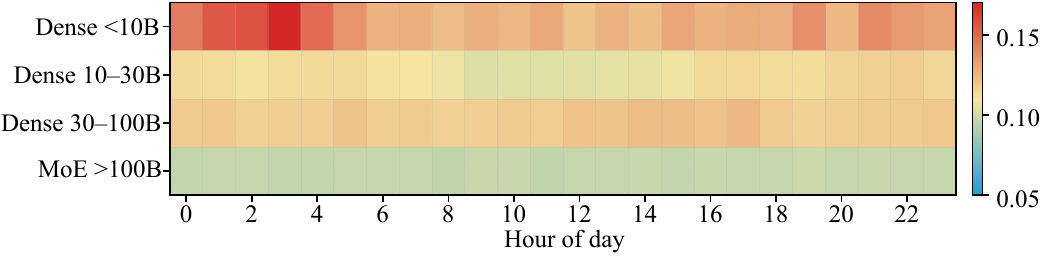}
    \caption{Top 5\% burst contribution by hour.}
    \label{fig:bursrtiness_con}
\end{figure}

Beyond characterizing local burstiness and fluctuation patterns, we further quantify the \emph{system-level importance} of bursty arrivals by measuring how much traffic mass is concentrated in extreme spikes. 
As shown in Fig.~\ref{fig:bursrtiness_con}, across all four architecture-scale groups, extreme seconds contribute a disproportionately large share of hourly traffic: even for the least burst-dominant category, the top 5\% seconds account for nearly 9.5\% of total requests, underscoring the non-negligible role of short-lived bursts in shaping aggregate load. 
This effect varies across model groups. Dense models, especially those under 10B parameters, exhibit the highest burst contribution, where the top 5\% seconds can account for up to 15--18\% of hourly requests during certain periods. This observation is consistent with our earlier finding that Dense workloads are dominated by frequent short-term fluctuations.

\begin{tcolorbox}[
colback=gray!10,
colframe=white,
boxrule=0pt,
arc=0pt,
left=3pt,
right=3pt,
top=3pt,
bottom=3pt
]
\textbf{Finding 2:} Short-term burstiness and fluctuations are multi-dimensional and architecture-dependent, with Dense models dominated by high-frequency fluctuations and low-amplitude bursts, while MoE models are characterized by lower-frequency, high-amplitude bursts.
\end{tcolorbox}


\subsubsection{Request Arrival Modeling}

Our analysis shows that LLM request arrivals are both \emph{non-stationary} and \emph{bursty}, and these properties vary systematically across model architectures and scales.
At coarse time scales, arrival intensity and distribution shift over time; at fine time scales, requests cluster within short intervals, leading to over-dispersion and pronounced bursts.
These observations motivate an architecture- and scale-aware, multi-level arrival modeling for credible benchmarking of multi-model serving systems.

Prior work such as BurstGPT models request arrivals by fitting a single inter-arrival distribution within fixed-length windows (e.g., 20 minutes), which captures coarse-grained arrival rates but overlooks short-term burstiness at fine time scales. To address this limitation, FineServe adopts a piecewise arrival modeling strategy that combines Gamma-based  modeling~\cite{AlpaServe, fastServe} with Negative Binomial (NB) modeling~\cite{nb_function, nb_function2} of fine-grained request counts.


We partition the request trace into fixed-length windows of $T=300$ seconds,
extract inter-arrival time (IAT) samples $\{x_i^{(s)}\}$ within each window, and fit their distribution with a Gamma model:
\begin{equation}
x^{(s)} \sim \mathrm{Gamma}(\alpha_s, \beta_s),
\end{equation}
where $\alpha_s$ and $\beta_s$ denote shape and scale parameters for window $s$.

\begin{figure}[htbp]
    \centering
    \includegraphics[width=\columnwidth]{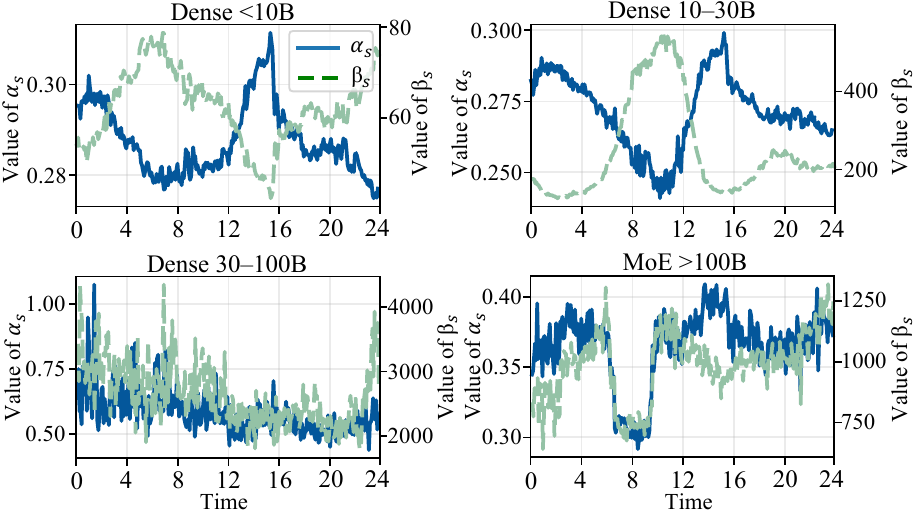}
    \caption{Temporal evolution of Gamma arrival parameters.}
    \label{fig:gamma}
\end{figure}



Fig.~\ref{fig:gamma} reports the temporal evolution of the fitted Gamma parameters, which implies the key arrival behaviors: the shape parameter $\alpha_s$ captures arrival regularity, whereas the scale parameter $\beta_s$ reflects request sparsity. Across architecture-scale groups, the fitted parameters differ systematically, indicating distinct usage patterns. 
Specifically, small Dense models exhibit relatively low $\alpha_s$ with small $\beta_s$, consistent with frequent and loosely structured arrivals; larger Dense models show higher $\alpha_s$ and much larger $\beta_s$, implying sparser yet more regular arrivals. MoE models fall in between, with moderate sparsity and pronounced temporal non-stationarity, consistent with more workload-driven invocation patterns in production.



Gamma-based inter-arrival modeling captures segment-level request timing but fails to represent fine-grained clustering when multiple requests arrive within the same instant. To check whether such clustering matters, we analyze arrival counts at millisecond granularity. As shown in Table~\ref{tab:nb} in the Appendix~\ref{ap:arrival_ct}, only the \texttt{Dense <10B} group exhibits clear over-dispersion, with a variance-to-mean ratio $>1$, indicating short-term request aggregation, while other model groups are too sparse at the millisecond scale, where fine-grained burstiness is negligible.

Accordingly, FineServe adopts a \emph{conditional} strategy. For architecture-scale groups with over-dispersed millisecond-level counts, we discretize time into $\Delta{=}1$\,ms slots and model the per-slot arrival count $n_{s,t}$ within window $s$ using an NB distribution:
\begin{equation}
n_{s,t} \sim \mathrm{NB}(r_s, p_s),
\end{equation}
where $r_s$ and $p_s$ control arrival dispersion and intensity, respectively. This formulation naturally allows multiple requests to arrive within the same time slot and captures short-term bursty aggregation.


\begin{tcolorbox}[
colback=gray!10,
colframe=white,
boxrule=0pt,
arc=0pt,
left=3pt,
right=3pt,
top=3pt,
bottom=3pt
]
\textbf{Finding 3:} In heterogeneous multi-model serving, modeling request arrivals with Gamma-distributed IAT alone is insufficient, accurately reproducing short-term concurrency requires additionally modeling millisecond-scale arrival counts (e.g., with an NB model).
\end{tcolorbox}

\subsubsection{Implications for LLM Serving and Elastic Cloud Systems}

\textbf{Long-term distribution shifts.} Arrival distributions exhibit pronounced shifts over days to months, invalidating stationary assumptions and necessitating elastic policies that adapt provisioning and routing as demand evolves. On multi-model platforms, these shifts are architecture- and scale-dependent. Treating all models as experiencing identical drift can misallocate capacity, causing under-provisioning of volatile small Dense groups with resulting overload and tail-latency spikes, while over-provisioning comparatively stable large MoE groups and wasting capacity.

\textbf{Short-term burstiness.} Burstiness is inherently multi-dimensional. Dense workloads are dominated by high-frequency second-level jitter, whereas MoE workloads exhibit lower-frequency, high-amplitude bursts that concentrate load within short intervals. A single variability metric is therefore insufficient: focusing only on burst magnitude misses Dense jitter, while focusing only on jitter misses MoE spikes. These differences suggest architecture-aware adaptation: fast-reacting, jitter-aware scheduling for Dense models and capacity-aware, burst-resilient provisioning for MoE models.


\subsection{Input-output Token Geometry}\label{sec:arch_scale_token}
\label{sec:token_geometry}

\begin{figure}[t]
    \centering
    \includegraphics[width=\columnwidth]{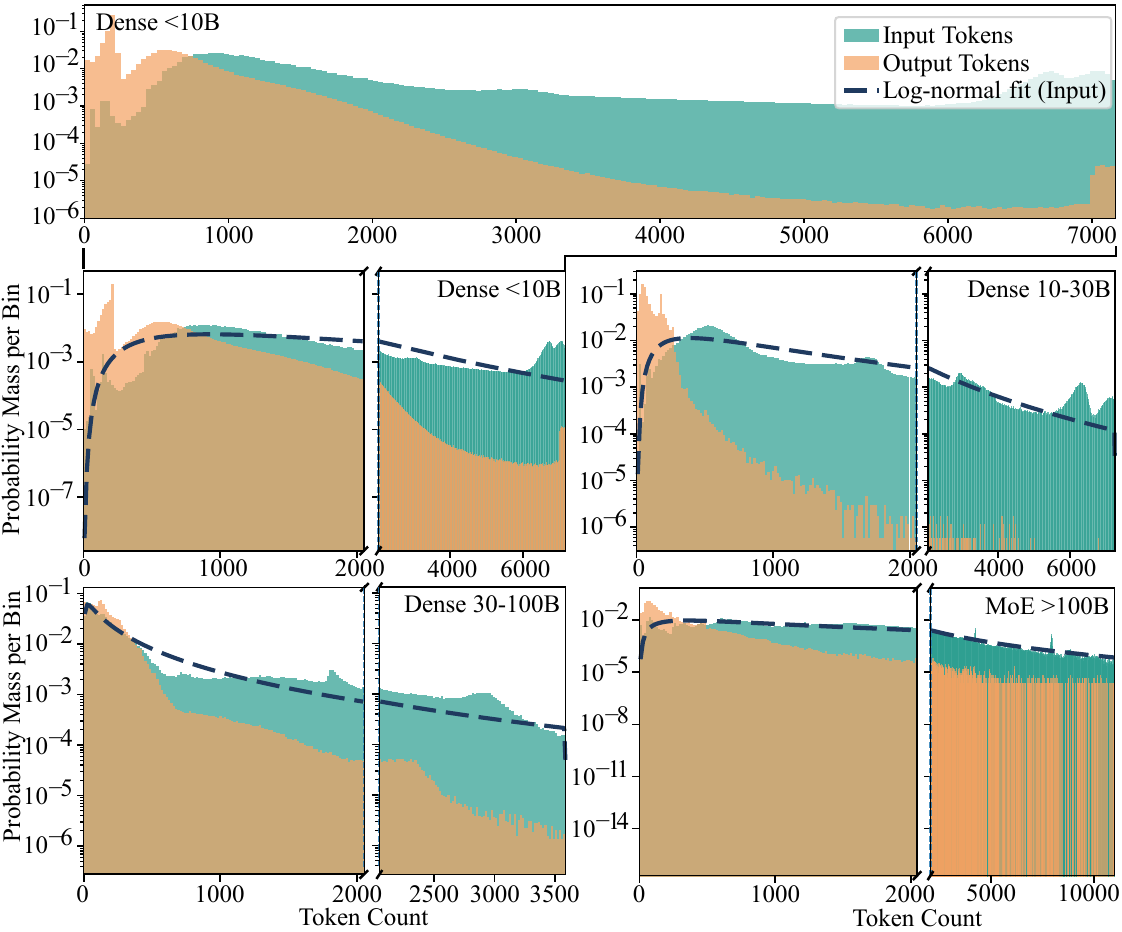}
    \caption{ Input and output length marginal distribution.}
    \label{fig:token_fre}
    \vspace{-0.5em}
\end{figure}

Request arrivals determine when workload enters the system, but token geometry determines how much computation each request requires. We characterize input and output token lengths across architecture-scale groups, examining both their marginal distributions and their conditional relationships. 

\subsubsection{Marginal Distributions}
Fig.~\ref{fig:token_fre} presents token-length histograms as probability mass per bin, well-suited to the heavy-tailed distributions observed in real-world workloads. Across all architecture-scale groups, input lengths exhibit substantially extended tails compared to those reported in BurstGPT, indicating that long-context prompts have become increasingly common. In contrast, output lengths decay more rapidly in the tail, although long generations remain non-negligible. Dense models tend to produce more concentrated outputs, whereas MoE models exhibit visibly heavier output tails. To enable scalable workload synthesis, we explicitly model input token length distributions. Unlike Zipf assumptions used in BurstGPT, our empirical data exhibit stable unimodal structures with smoother tail decay on the logarithmic scale, motivating log-normal input length modeling: 
\vspace{-0.2em}
\begin{equation}
\label{eq:input_lognormal}
P(\text{Input}) = \text{Log-normal}(\mu, \sigma^2),
\vspace{-0.2em}
\end{equation}
where $\mu$ and $\sigma$ denote the mean and standard deviation of the logarithm of input token lengths.


\begin{tcolorbox}[
colback=gray!10,
colframe=white,
boxrule=0pt,
arc=0pt,
left=3pt,
right=3pt,
top=3pt,
bottom=3pt
]
\textbf{Finding 4:} Token distributions reveal a clear transition toward long-context regimes. Input tails extend substantially beyond prior reports, while MoE models exhibit heavier output tails than Dense models. Log-normal distributions capture both central mass and tail behavior more faithfully than Zipf assumptions.
\end{tcolorbox}

\begin{figure}[bp]
    \centering
    \includegraphics[width=\columnwidth]{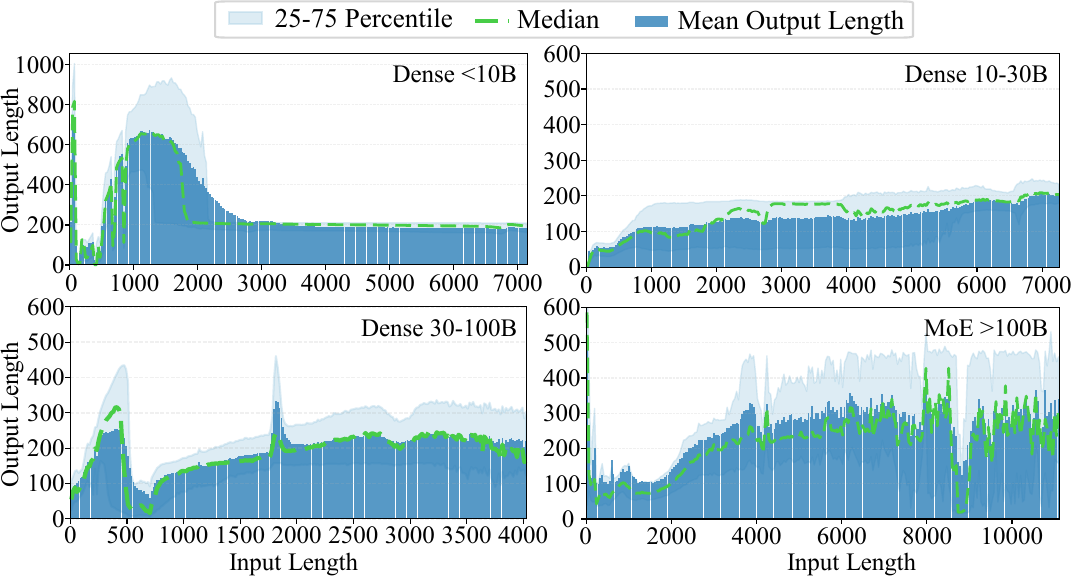}
    \caption{Input-output correlation across model categories.}
    \label{fig:token_model_relation}
\end{figure}

\subsubsection{Input-Output Correlation}
While marginal distributions characterize overall statistics, they cannot capture dependencies between input and output lengths. Fig.~\ref{fig:token_model_relation} reveals architecture-specific patterns through median output trajectories across input bins.

Contrary to simple monotonic assumptions, Dense models, particularly those below 10B parameters, exhibit a pronounced non-monotonic, inverted-bowl relationship between input and output lengths. As input length increases, the median output grows from short contexts to a peak at moderate inputs around 1.5K–2K tokens, and then declines substantially for longer contexts, stabilizing at much shorter outputs for very long prompts. From the peak to the long-input regime, the median output length drops by more than 60\%. Larger Dense models show similar but attenuated trends: Dense 30–100B models exhibit a weaker inverted-bowl pattern with lower peak outputs, while Dense 10–30B models largely approach near-monotonic behavior.

This behavior may reflect a combination of workload semantics and architectural effects. As input length increases, requests increasingly correspond to summarization or condensation scenarios that favor concise outputs, whereas moderate-length inputs more often elicit generative tasks requiring expanded responses. At the same time, Dense architectures appear to compress information from long contexts through full-parameter attention, distilling salient content rather than expanding outputs proportionally. In contrast to the roughly linear input–output correlation reported in BurstGPT, our observations suggest that input–output geometry is strongly shaped by model architecture and scale.


By contrast, MoE models exhibit a monotonic growth–saturation pattern. Median output length increases steadily with input length and saturates at high-context regimes, reflecting MoE’s sparse expert routing that allows longer inputs to activate more diverse expert combinations rather than being aggressively compressed. This architectural contrast leads to a crossover effect: Dense models produce longer outputs at moderate input lengths, while MoE models dominate in long-context regimes. Within Dense models, parameter scale introduces secondary modulation. 

\begin{tcolorbox}[
colback=gray!10,
colframe=white,
boxrule=0pt,
arc=0pt,
left=3pt,
right=3pt,
top=3pt,
bottom=3pt
]
\textbf{Finding 5:} Input-output relationships are fundamentally architecture-dependent and non-monotonic. Dense models exhibit inverted-bowl patterns (peak at moderate inputs, substantial decline for long contexts), while MoE models show sustained growth (3.5--4$\times$) with saturation. 
\end{tcolorbox}

\section{Characterizing Tasks}

Different task intents pursue different end goals and require different types and amounts of information, which can lead to systematically distinct input and output characteristics and, in turn, shape system load and resource demand. Motivated by this observation, we conduct a fine-grained analysis of task composition and task-specific input and output token properties. Unlike prior studies~\cite{stateofai} that primarily report task proportions, we focus on how task intent influences token geometry, providing insights that are directly relevant to LLM system design and evaluation.

We perform task classification on a random sample of 100K requests from our dataset. Requests are mapped into 10 representative task intent categories (detailed in the Appendix~\ref{ap:models_tasks}). To reflect multi-intent prompts, we retain up to two high-confidence task labels per request. All classification is executed offline on sanitized prompt content in a privacy-preserving manner.

The classification pipeline proceeds in two stages. In the first stage, following prior work~\cite{stateofai}, we apply Google Cloud Natural Language's \texttt{classifyText} API to a labeled subset of 20K requests drawn from internal user interactions, which are pre-approved for research use. The API maps textual input (up to the first 1,000 characters of each prompt) to a hierarchical, language-agnostic taxonomy, returning category paths with associated confidence scores in [0,1]; categories below the default confidence threshold of 0.5 are discarded. We then consolidate the API's fine-grained taxonomy into our 10 
study-defined task intent buckets, yielding 20K silver-labeled 
annotated requests as a training corpus.

In the second stage, we use this labeled corpus to fine-tune a BERT-based classifier~\cite{bert} for multi-label classification, retaining up to two predicted labels per request. The model is trained and served entirely within our internal server cluster, so that no prompt content is transmitted to external services during inference. To further protect user privacy, predictions are logged at the request level without any linkage to individual users. The fine-tuned model is then applied in batch to the full 100K-request sample in an automated offline pipeline, producing the task-intent labels used in all subsequent analyses.

\begin{figure}[htbp]
    \centering
    \includegraphics[width=0.9\columnwidth]{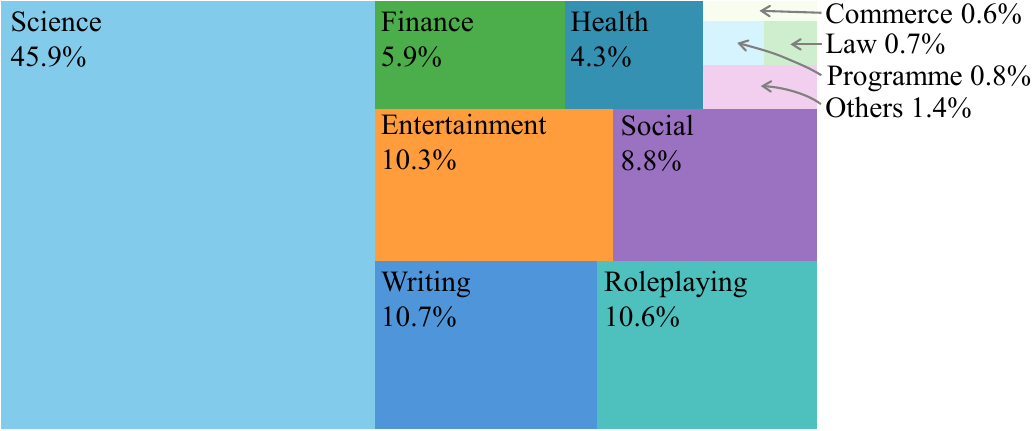}
    \caption{Distribution of task intents.}
    \label{fig:task_treemap}
\end{figure}

\begin{figure}[b]
    \centering
    \includegraphics[width=\columnwidth]{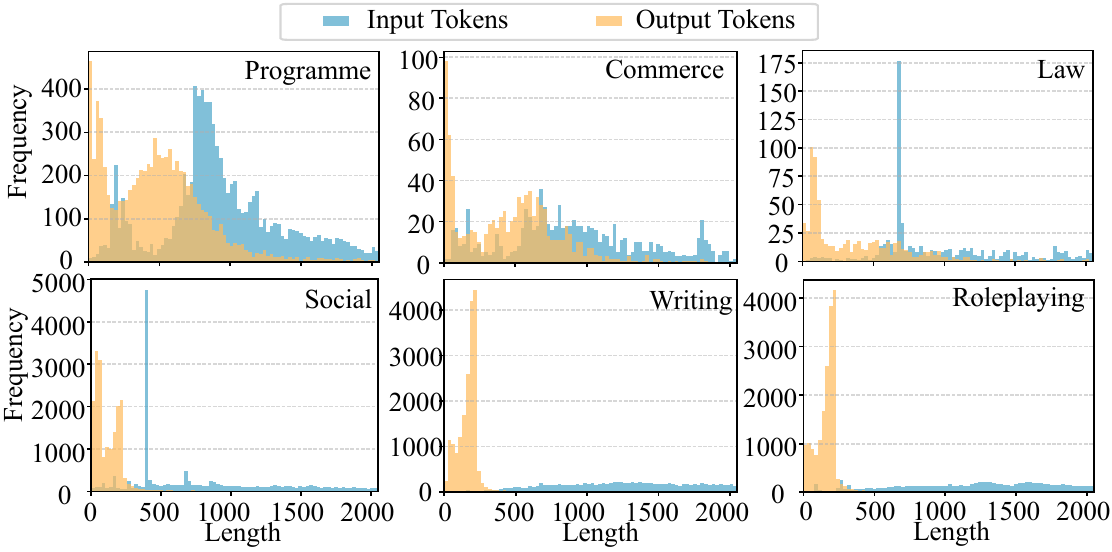}
    \caption{Task-specific marginal token distributions.}
    \label{fig:task_marginal}
\end{figure}

\subsection{Task Composition}
In practice, deployed LLM services rarely operate under single-task settings, but instead handle a mixture of requests spanning multiple application domains. Characterizing task composition therefore provides essential context for understanding workload heterogeneity and its implications for system behavior.

Fig.~\ref{fig:task_treemap} shows the distribution of task intents in our dataset. Scientific requests account for the largest share, comprising nearly half of all requests. Writing, role-playing, and entertainment tasks also contribute substantial fractions, while categories such as law, programming, and commerce appear less frequently and exhibit long-tailed behavior. Workload displays a diverse and heterogeneous task composition rather than being dominated by an intent, reflecting realistic usage patterns in production LLM services.

\subsection{Task-specific Token Characteristics}\label{sec:task_token}
Complementary to the architecture-scale analysis in \S\ref{sec:arch_scale_token}, we characterize task-driven heterogeneity in token-length patterns. We analyze both marginal distributions of input and output lengths and conditional relationships between them under different task intents, providing quantitative foundations for task-aware workload generation and downstream system design, including buffer allocation and prefill-decode resource planning.

\begin{figure}[htb]
    \centering
    \includegraphics[width=\columnwidth]{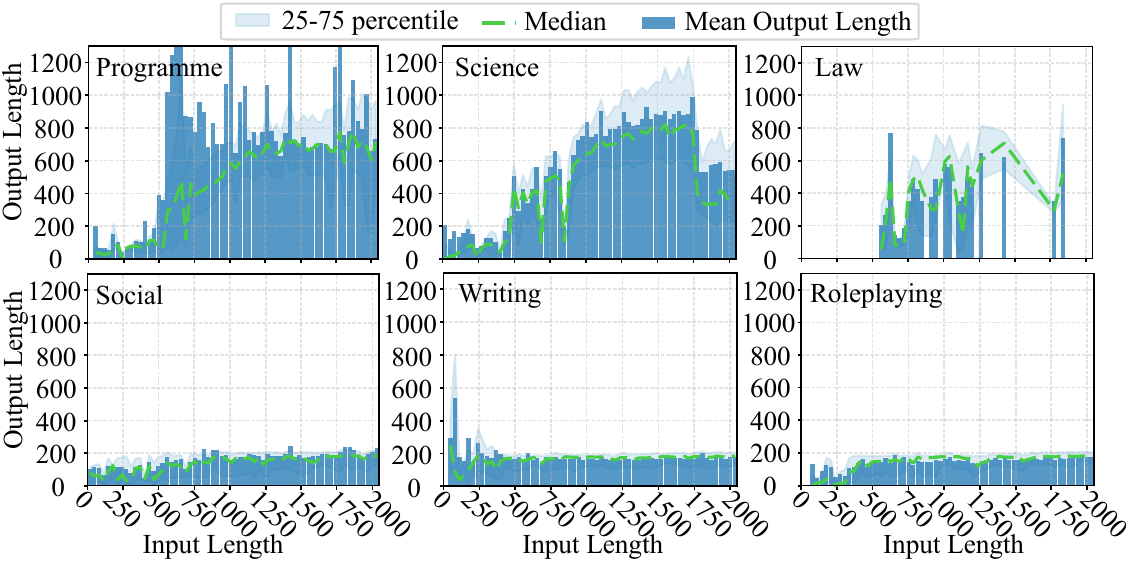}
    \caption{Task-specific input-output correlation patterns.}
    \label{fig:task_io}
\end{figure}

\subsubsection{Marginal Distributions}
Fig.~\ref{fig:task_marginal} shows input-token histograms for major task categories.
Unlike architecture-specific patterns that differ primarily in scale and tail weight, task heterogeneity manifests through distribution shapes. Some tasks concentrate mass in continuous humps with smooth tails (Programme, Commerce), others exhibit needle-like spikes at stable locations plus low-density backgrounds (Law, Science, Finance, Health, Social), and a third group spreads mass across wide plateaus with no dominant mode (Writing, Entertainment, Role-playing). This shape-driven taxonomy remains stable across random subsampling and supports generator-friendly modeling. In contrast, output distributions remain largely consistent across tasks, with most mass concentrated in 0--300 tokens and only minor tail variations. This consistency implies that task-awareness is most critical for modeling prefill-side input load, while decode-side output statistics admit a shared base distribution with light task-specific corrections.



\begin{tcolorbox}[
colback=gray!10,
colframe=white,
boxrule=0pt,
arc=0pt,
left=3pt,
right=3pt,
top=3pt,
bottom=3pt
]
\textbf{Finding 6:} Task intents induce strong heterogeneity in input distributions through shape-level differences (hump-tail, spike-background, wide-plateau), while output distributions remain largely consistent. Task-aware generation must explicitly model prefill-side diversity, while decode-side can rely on shared parameterizations.
\end{tcolorbox}

\subsubsection{Input-Output Correlation}
Fig.~\ref{fig:task_io} shows median output trajectories across input bins for representative tasks (the complete set of ten tasks is provided in Appendix~\ref{full_task}). Technical and knowledge-intensive tasks (Programme, Science, Finance, Health, Commerce, Law) exhibit rise–peak–decline–stable patterns that closely resemble the Dense inverted-bowl geometry observed in \S\ref{sec:arch_scale_token}. Median output increases with input length at moderate scales and then declines toward the tail regime. This suggests a semantic shift: moderate contexts trigger expanded explanations or step-by-step reasoning, while very long contexts correspond to compression-oriented tasks such as summarization.

In contrast, conversational and creative tasks (Social, Writing, Entertainment, Role-playing) are largely input-insensitive, with median output remaining approximately constant across wide input ranges and tight percentile bands. This suggests that high-volume interactive tasks can be served using conservative decode buffers.

\begin{tcolorbox}[
colback=gray!10,
colframe=white,
boxrule=0pt,
arc=0pt,
left=3pt,
right=3pt,
top=3pt,
bottom=3pt
]
\textbf{Finding 7:} Task patterns exhibit structural compatibility with architecture patterns. Technical tasks show inverted-bowl profiles (peak at moderate inputs, decline for long contexts), while conversational tasks remain flat. This cross-dimensional consistency enables compositional generation combining architecture-aware and task-aware modules.
\end{tcolorbox}

\section{Workload Generation}\label{generator}



\subsection{FineServe Framework}
\begin{figure}[tbp]
    \centering
    \includegraphics[width=\columnwidth]{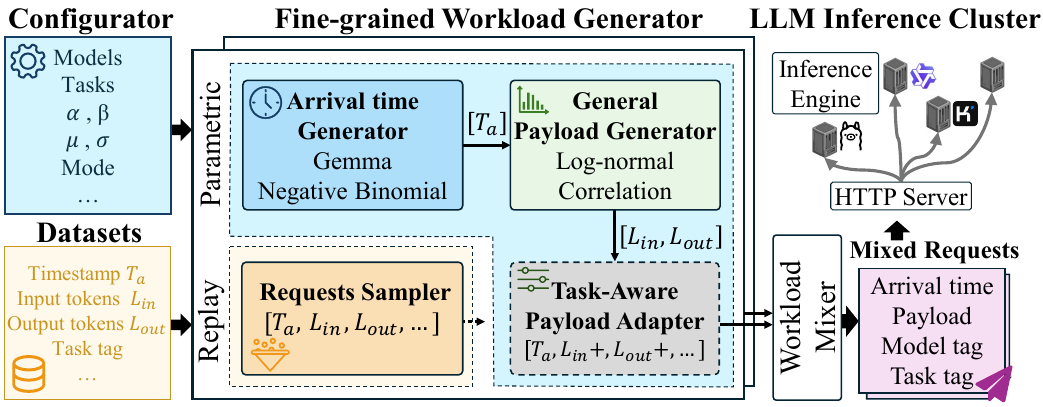}
    \caption{Overview of the FineServe framework. Per-model request streams are mixed to produce a unified multi-model workload benchmark. The color gray indicates task-aware payload adaptation is optional, allowing FineServe to naturally degenerate to general task workload simulation. }
    \label{fig:overview}
\end{figure}

FineServe provides a lightweight and practical framework for generating realistic multi-model LLM serving workloads (Fig.~\ref{fig:overview}).
At a high level, FineServe constructs per-model request streams and then aggregates them into a unified mixed workload for evaluation in an inference cluster.
To use FineServe, a user first specifies the target model set, the task composition (optional), and the desired workload scale through the \emph{Configurator}.
FineServe then invokes the \emph{Fine-grained Workload Generator} to produce each per-model stream in one of two modes: (i) \emph{real-trace replay}, which directly samples requests from the trace repository while preserving their original timestamps and payloads, and (ii) \emph{parametric synthesis}, which samples timestamps and payloads from fitted statistical models to scale workloads to arbitrary volumes.

In parametric synthesis, FineServe follows an \emph{arrival-then-payload} pipeline.
It first samples request timestamps using model-aware arrival models (Gamma inter-arrivals, augmented with NB count modeling when millisecond-level concurrency is present), capturing both long-term rate shifts and short-term burstiness.
It then samples correlated input/output token lengths using class-specific payload models, optionally refining payloads via a task-aware adapter to match the configured task composition.
Finally, a request sampler assembles $(T_a, L_{\text{in}}, L_{\text{out}}, \text{model}, \text{task})$ tuples for each model, where $T_a$ is the arrival time, $(L_{\text{in}}, L_{\text{out}})$ denotes input/output token lengths (and associated prompt metadata from ShareGPT or other public datasets). Then, the \emph{Workload Mixer} merges per-model streams into a single multi-model workload.
Overall, FineServe operationalizes the workload signatures identified in earlier sections and exposes them as configurable knobs, enabling scalable yet realistic workload construction for benchmarking multi-model LLM serving systems.

\subsection{Parametric Generation vs. Trace Replay}

\begin{figure}[tbp]
    \centering
    \includegraphics[width=\columnwidth]{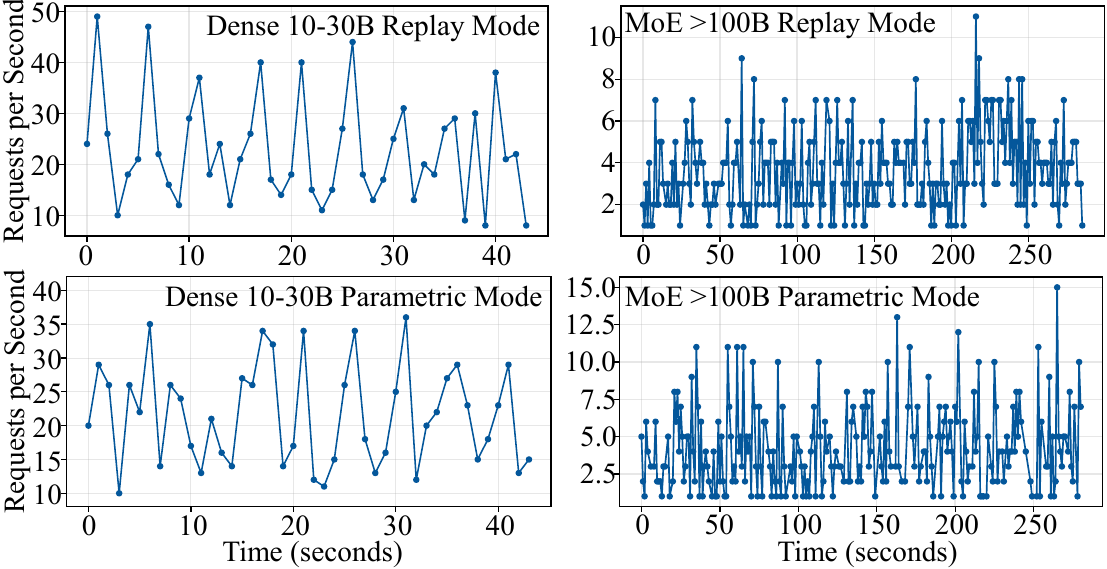}
    \caption{Request rate over time generated by replay and parametric modes. }
    \label{fig:use}
\end{figure}

To validate the fidelity of our parametric workload generation, we compare the arrival patterns produced by the parametric mode against those obtained via trace replay. For each architecture group, we select a parameter scale and generate 1,000 requests using the default fitted parameters. We then plot the resulting per-second request rates alongside their replayed counterparts.

As shown in Fig.~\ref{fig:use}, the parametric workloads closely resemble the replayed traces in terms of arrival intensity, burst frequency, and temporal fluctuation patterns, across both Dense and MoE model families. We emphasize that our goal is not to reproduce the original arrival sequence verbatim, but rather to approximate its empirical arrival dynamics at a macroscopic timescale. This parameterized formulation enables researchers to flexibly generate workloads with controllable scale, burstiness, and variability, thereby facilitating reproducible evaluation and scheduling studies under diverse operating regimes.


\section{Conclusion}\label{conclusion}
We present a comprehensive and fine-grained characterization of real-world LLM serving workloads for today’s multi-model platforms. We analyze workload heterogeneity along model architecture/scale tiers and task intents, and unveil various meaningful findings. 
Motivated by these findings, we introduce FineServe, a model-aware workload generation framework that supports both real-trace replay and parametric synthesis, enabling realistic multi-model workload mixing and scalable evaluation. 
We believe that the dataset, analyses, and generation framework can support more faithful benchmarking and system studies on LLM serving systems.


\bibliographystyle{ACM-Reference-Format}
\bibliography{sample-base}

\clearpage
\appendix



\section{Per-Millisecond Arrival Count Statistics}\label{ap:arrival_ct}
\begin{table}[htbp]
  \centering
  \setlength{\tabcolsep}{3 pt}
  \caption{Per-millisecond arrival count statistics and architecture-aware modeling decisions.}
    \begin{tabular}{c|ccccc}
    \toprule
    Category & Mean & Var & Var/Mean & NB $r_s$ & Function \\
    \toprule
    \texttt{Dense <10B} & 2.63  & 2.97  & 1.13  & 0.03696 & Gamma+NB \\
    \texttt{Dense 10–30B} & 1.12  & 0.12  & 0.11  & 0.00002 & Gamma-only \\
    \texttt{Dense 30–100B} & 1     & 0.003 & 0.003 & 0.00002 & Gamma-only \\
    \texttt{MoE >100B} & 1.01  & 0.01  & 0.01  & 0.00002 & Gamma-only \\
    \bottomrule
    \end{tabular}%
  \label{tab:nb}%
\end{table}%

\section{Full Task-Specific Token Distributions} \label{full_task}

\begin{figure}[htbp]
    \centering
    \includegraphics[width=\columnwidth]{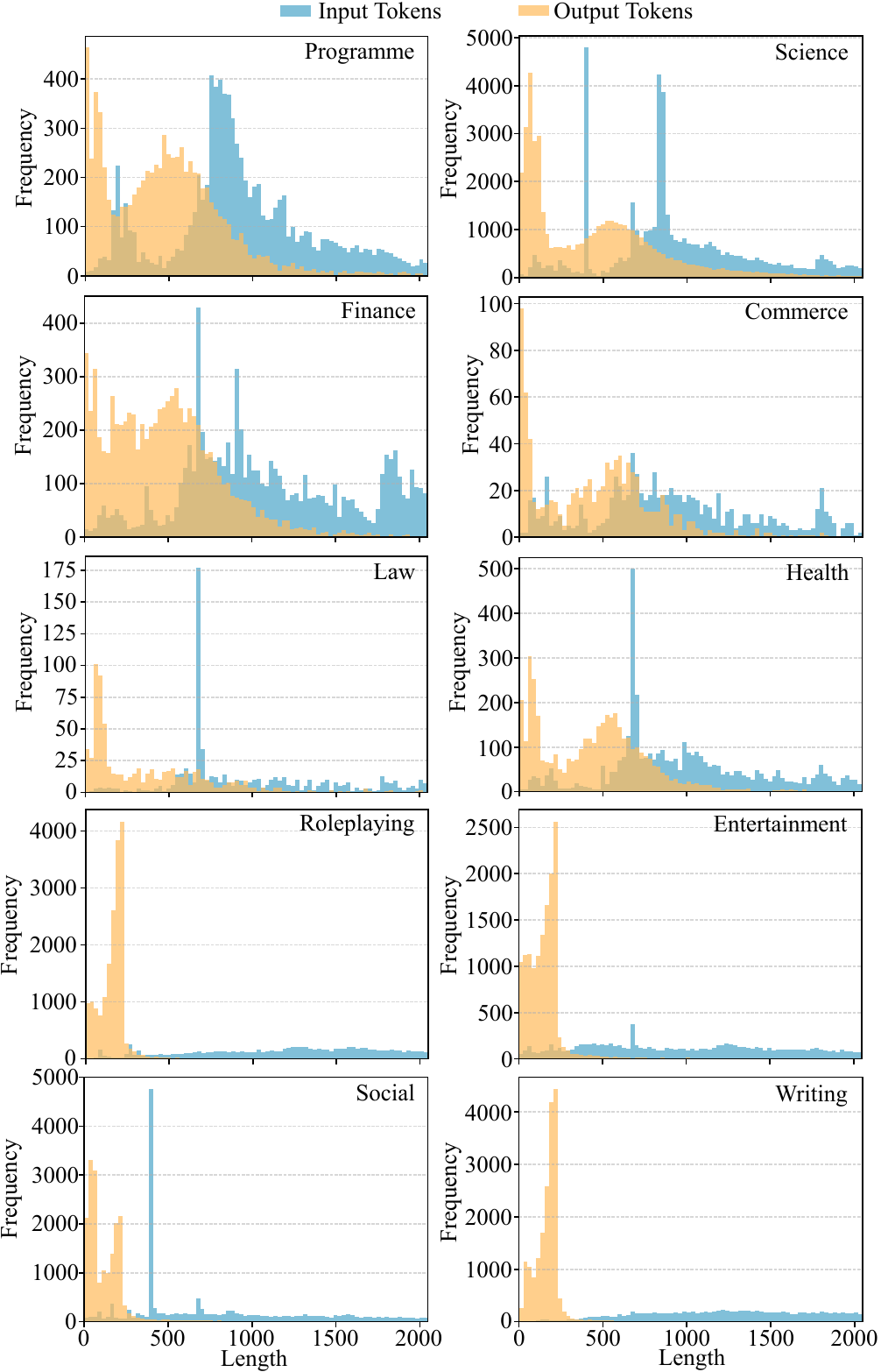}
    \caption{Distributions of input and output token lengths across tasks. For each task intent, the figure shows the empirical distributions of input tokens and output tokens, truncated at 2048 tokens for visualization.}
    \label{fig:count}
\end{figure}

\begin{figure}[htbp]
    \centering
    \includegraphics[width=\columnwidth]{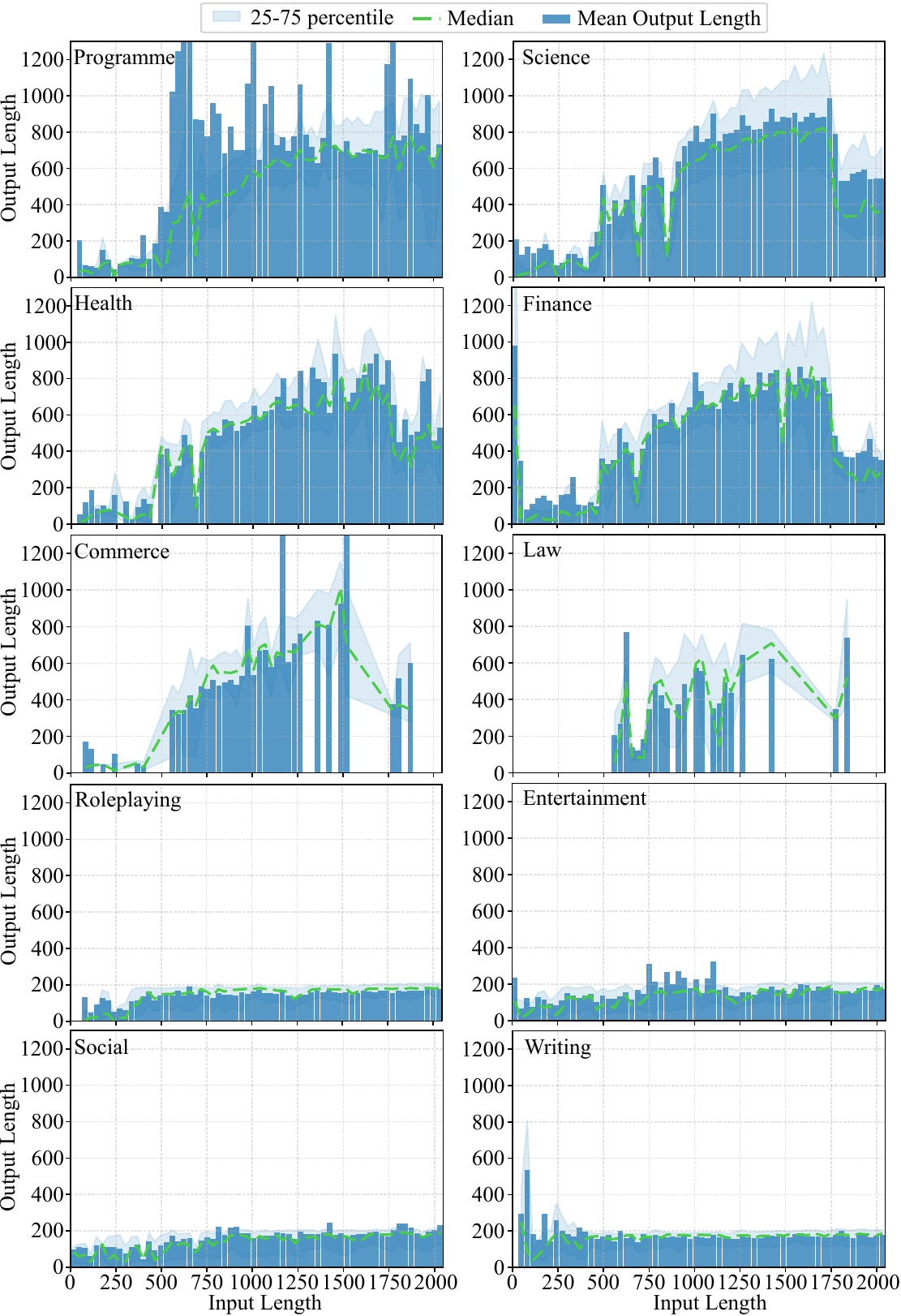}
    \caption{Input–output token length relationships across tasks.
For each task intent, the figure reports conditional statistics of output token length as a function of input (prompt) length, including the mean (bars), median (dashed line), and interquartile range (25–75 percentile, shaded).}
    \label{fig:relation}
\end{figure}

Fig.~\ref{fig:count} and Fig.~\ref{fig:relation} present the complete input and output token length distributions for all ten task intent categories, complementing the representative results discussed in the main text.

\section{Parametric Modeling}
\label{sec:appendix_parametric_model}
\subsection{Architecture-Specific Input--Output Token Models}
\label{sec:appendix_arch_io}
Let $x$ and $y$ denote the input and output token lengths of a request, respectively.
We characterize Dense outputs via non-monotonic piecewise functions, where each segment corresponds to a distinct operational regime:
\begin{equation}
\label{eq:dense_io}
\mathbb{E}[y | x]_{\text{Dense}} \approx \begin{cases}
O_{\text{base}} + \alpha_{\text{Dense}} \cdot x, & x < \tau_{\text{peak}}, \\
O_{\text{peak}} - \beta_{\text{Dense}} \cdot (x - \tau_{\text{peak}}), & \tau_{\text{peak}} \leq x < \tau_{\text{stable}}, \\
O_{\text{tail}}, & x \geq \tau_{\text{stable}}.
\end{cases}
\end{equation}
The rising phase is characterized by slope $\alpha_{\text{Dense}}$, reflecting context accumulation that peaks at input length $\tau_{\text{peak}}$ with output $O_{\text{peak}}$. The declining phase with slope $\beta_{\text{Dense}}$ captures compression-task dominance, stabilizing beyond $\tau_{\text{stable}}$ at output level $O_{\text{tail}}$. 
We sample the residual as $\varepsilon \sim \mathcal{N}(0,\sigma_{\text{Dense}}(x))$ to match empirical percentile bands.
This formulation enables input-aware buffer allocation strategies.

We model MoE outputs as:
\begin{equation}
\label{eq:moe_io}
\mathbb{E}[y | x]_{\text{MoE}} \approx \begin{cases}
O_{\text{init}} + k_1 \cdot x, & x < \tau_{\text{growth}}, \\
O_{\text{mid}} + k_2 \cdot (x - \tau_{\text{growth}}), & \tau_{\text{growth}} \leq x < \tau_{\text{sat}}, \\
\min(O_{\text{sat}} + k_3 \cdot (x - \tau_{\text{sat}}), M_{\max}), & x \geq \tau_{\text{sat}},
\end{cases}
\end{equation}
where initial output $O_{\text{init}}$ grows with slope $k_1$ until $\tau_{\text{growth}}$, then accelerates with slope $k_2$ from $O_{\text{mid}}$ until saturation onset at $\tau_{\text{sat}}$, finally plateauing at ceiling $M_{\max}$ with residual slope $k_3$. 
$M_{\max}$ is the output-length ceiling (in tokens) in the saturated regime.
Table~\ref{tab:io_function_params} summarizes key parameters for both architectures.

\begin{table}[t]
\centering
\caption{Fitted parameters for input-output correlation modeling. Piecewise 
functions capture median trends across architectures.}
\label{tab:io_function_params}
\small
\begin{tabular}{llll}
\toprule
\textbf{Arc.} & \textbf{Function Type} & \textbf{Key Stages} & \textbf{Characteristics} \\
\midrule
\multirow{3}{*}{Dense} & Non-monotonic & Rise & Peak: 620 @ 1500 input \\
                       & Piecewise     & Decline & Stable: 165 @ 5000+ \\
                       &               & Stable & Drop: 60\% \\
\midrule
\multirow{3}{*}{MoE}   & Growth-       & Slow Growth & Start: 100 @ 500 \\
                       & Saturation    & Linear Growth & Mid: 290 @ 5000 \\
                       & Piecewise     & Saturation & Max: 400 @ 8000+ \\
\bottomrule
\end{tabular}
\vspace{-6pt}
\end{table}

 Generator implementations sample input from Eq.~\eqref{eq:input_lognormal}, compute median output via Eq.~\eqref{eq:dense_io} or~\eqref{eq:moe_io} using parameters from Table~\ref{tab:io_function_params}, and overlay variance from empirical P25--P75 bands. For serving systems, Dense workloads require input-aware buffer allocation (large buffers for moderate inputs, reduced allocation for long contexts), while MoE workloads require dynamic expansion strategies.

\subsection{Task-Specific Token Modeling}
\label{sec:appendix_task_model}

\subsubsection{Task-Specific Marginal Distributions}
\label{sec:appendix_task_marginal}
 We model task inputs using three families corresponding to the observed shape patterns. For Hump-Tail tasks (Programme, Commerce):
\begin{equation}
\label{eq:task_input_hump}
P_t(X) = \pi_t \cdot \text{LogNormal}(\mu_t, \sigma_t^2) + (1 - \pi_t) \cdot \text{Exp}(\lambda_t),
\end{equation}
where $\pi_t$ controls the weight of the core distribution, $(\mu_t, \sigma_t)$ parameterize the log-normal component, and $\lambda_t$ controls exponential tail decay.

For Spike-Background tasks (Law, Science, Finance, Health, Social):
\begin{equation}
\label{eq:task_input_spike}
P_t(X) = \sum_{k=1}^{K_t} \rho_{t,k} \cdot \delta(X - \tau_{t,k})
+ \left(1 - \sum_k \rho_{t,k}\right) \cdot q_t(X),
\end{equation}
where $\delta(\cdot)$ denotes point masses at spike locations $\{\tau_{t,k}\}$, $\rho_{t,k}$ are spike weights, and $q_t(X)$ is a continuous background distribution.

For Wide-Plateau tasks (Writing, Entertainment, Role-playing):
\begin{equation}
\label{eq:task_input_plateau}
P_t(X) \propto \mathbf{1}[X \in (a_t, b_t)],
\end{equation}
a truncated uniform over range $(a_t, b_t)$.

All tasks share a common output base:
\begin{equation}
\label{eq:task_output_base}
P(Y) = \text{LogNormal}(\mu_{\text{out}}, \sigma_{\text{out}}^2).
\end{equation}

\subsubsection{Task-Conditional Input--Output Modeling}
\label{sec:appendix_task_conditional}
We model task-conditional outputs via a two-component framework. For task $t$:
\begin{equation}
\label{eq:task_conditional_general}
y = g_{h(t)}(x; \theta_t) + \varepsilon, \quad \varepsilon \sim \mathcal{N}(0, \sigma_t^2(x)),
\end{equation}
where $h(t)$ maps task $t$ to its pattern class (inverted-bowl or flat), $g_{h(t)}(\cdot)$ is the class-specific median function, $\varepsilon$ captures residual variability and $\sigma_t^2(x)$ is the variance.

For inverted-bowl tasks (technical and knowledge-intensive):
\begin{equation}
\label{eq:task_io_inv}
g_{\text{inv}}(x; \theta_t) = \begin{cases}
\alpha_t x, & x < \tau^{(t)}_{\text{peak}}, \\
O^{(t)}_{\text{peak}} - \beta_t(x - \tau^{(t)}_{\text{peak}}), & \tau^{(t)}_{\text{peak}} \leq x < \tau^{(t)}_{\text{stable}}, \\
O^{(t)}_{\text{tail}}, & x \geq \tau^{(t)}_{\text{stable}}.
\end{cases}
\end{equation}

For flat tasks (conversational and creative):
\begin{equation}
\label{eq:task_io_flat}
g_{\text{flat}}(x; \theta_t) = c_t,
\end{equation}
where $c_t$ is a task-specific constant.

These findings inform both workload generation and serving system design. Generator implementations classify tasks into pattern classes, apply corresponding median functions (Eq.~\eqref{eq:task_io_inv} or~\eqref{eq:task_io_flat}) as shown in Table~\ref{tab:task_io_params}, and overlay variance sampled from task-specific percentile bands. For serving systems, task-aware routing enables technical tasks to benefit from input-aware buffer allocation (large buffers for moderate inputs, reduced allocation for long contexts), while conversational tasks use fixed conservative buffers that simplify memory management.

\begin{table}[h]
\centering
\caption{Task-specific input-output correlation parameters. Two pattern classes capture median trends across task intents.}
\label{tab:task_io_params}
\small
\begin{tabular}{llccc}
\toprule
\textbf{Pattern} & \textbf{Example Tasks} & $\tau_{\text{peak}}$ & $O_{\text{peak}}$ & $O_{\text{tail}}$ \\
\midrule
Inverted-bowl & Programme & 1100 & 950 & 520 \\
 & Science & 1200 & 860 & 540 \\
 & Law & 750 & 620 & 280 \\
\midrule
Flat & Social & -- & -- & 140 (const) \\
 & Writing & -- & -- & 155 (const) \\
\bottomrule
\end{tabular}
\end{table}

\begin{table*}[htbp]
  \centering
  \small
  \caption{Model suite used in our analysis. Models are grouped by architecture (Dense vs. MoE) and parameter scale (\texttt{<10B, 10--30B, 30--100B, >100B}), and annotated with their public release month to contextualize model recency.}
  \begin{tabular}{c|c|ccc}
    \toprule
    Architecture & Scale & Model & Model Series & Release \\
    \midrule
    \multirow{42}{*}{Dense}
    & \multirow{19}{*}{\texttt{<10B}}
      & Qwen3-4B-FP8 & Qwen & Jan-25 \\
    & & Qwen3-8B-FP8 & Qwen & Jan-25 \\
    & & Qwen2.5-7B-Instruct & Qwen & Oct-24 \\
    & & DeepSeek-R1-0528-Qwen3-8B & DeepSeek & Jun-24 \\
    & & Mistral-7B-Instruct & Mistral & Oct-23 \\
    & & LLaMA-3.2-1B-Instruct & LLaMA & Sep-24 \\
    & & LLaMA-3.2-3B-Instruct & LLaMA & Sep-24 \\
    & & LLaMA-3.1-8B-Instruct-BF16 & LLaMA & Jul-24 \\
    & & LLaMA-3.1-8B-Instruct-FP8 & LLaMA & Jul-24 \\
    & & LLaMA-3-8B-Instruct & LLaMA & Apr-24 \\
    & & Hermes-2-Pro-LLaMA-3-8B & LLaMA & Jun-24 \\
    & & DeepSeek-R1-Distill-LLaMA-8B & DeepSeek & Jun-24 \\
    & & GLM-Z1-9B-0414 & GLM & Apr-24 \\
    & & GLM-4-9B-0414 & GLM & Apr-24 \\
    & & L3-8B-Stheno-v3.2 & Sao10K & Nov-24 \\
    & & L3-8B-Lunaris & Sao10K & Nov-24 \\
    & & L3-8B-Stheno-v3.2-SpicyChat & Sao10K & Nov-24 \\
    & & WizardLM-2-7B & WizardLM & Jun-24 \\
\cmidrule{2-5}
    & \multirow{8}{*}{\texttt{10--30B}}
      & Gemma-3-27B-IT & Gemma & Dec-24 \\
    & & Mistral-Nemo & Mistral & Jul-24 \\
    & & DeepSeek-R1-Distill-Qwen-14B & DeepSeek & Jun-24 \\
    & & MythoMax-L2-13B & LLaMA & Dec-23 \\
    & & Captain-Eris\_Violet-v0.420-12B & LLaMA & Dec-24 \\
    & & MN-12B-Lyra-Spiced-v1-FP8 & LLaMA & Dec-24 \\
    & & MN-12B-Mag-Mell-R1-Yodayo & LLaMA & Dec-24 \\
\cmidrule{2-5}
    & \multirow{15}{*}{\texttt{30--100B}}
      & Qwen3-32B-FP8 & Qwen & Jan-25 \\
    & & Qwen2.5-72B-Instruct & Qwen & Oct-24 \\
    & & Qwen2.5-VL-72B-Instruct & Qwen & Oct-24 \\
    & & Dolphin-2.9.2-Qwen2-72B & Dolphin & Nov-24 \\
    & & DeepSeek-R1-Distill-Qwen-32B & DeepSeek & Jun-24 \\
    & & LLaMA-3.3-70B-Instruct & LLaMA & Dec-24 \\
    & & LLaMA-3-70B-Instruct & LLaMA & Apr-24 \\
    & & DeepSeek-R1-Distill-LLaMA-70B & DeepSeek & Jun-24 \\
    & & LLaMA-3.1-70B-Euryale-v2.2 & Sao10K & Dec-24 \\
    & & LLaMA-3-70B-Euryale-v2.1 & Sao10K & Nov-24 \\
    & & Midnight-Rose-70B & Sao10K & Dec-24 \\
    & & Midnight-Rose-70B-SpicyChat & Sao10K & Dec-24 \\
    & & GLM-Z1-32B-0414 & GLM & Apr-24 \\
    & & GLM-4-32B-0414 & GLM & Apr-24 \\
    & & GLM-Z1-Rumination-32B-0414 & GLM & Apr-24 \\
    \midrule
    \multirow{16}{*}{MoE}
    & \multirow{3}{*}{\texttt{10--30B}}
      & Qwen3-30B-A3B-FP8 & Qwen & Jan-25 \\
    & & LLaMA-4-Scout-17B-16E-Instruct & LLaMA & Jan-25 \\
    & & LLaMA-4-Maverick-17B-128E-Instruct-FP8 & LLaMA & Jan-25 \\
\cmidrule{2-5}
    & \multirow{12}{*}{\texttt{>100B}}
      & Qwen3-235B-A22B-FP8 & Qwen & Jan-25 \\
    & & DeepSeek-Prover-V2-671B & DeepSeek & Aug-24 \\
    & & DeepSeek-R1-Turbo & DeepSeek & Jul-24 \\
    & & DeepSeek-V3-Turbo & DeepSeek & Jun-24 \\
    & & DeepSeek-R1-0528 & DeepSeek & May-24 \\
    & & DeepSeek-R1 & DeepSeek & May-24 \\
    & & DeepSeek-R1-Community & DeepSeek & May-24 \\
    & & DeepSeek-V3-Community & DeepSeek & Apr-24 \\
    & & DeepSeek-V3-0324 & DeepSeek & Mar-24 \\
    & & DeepSeek-V3 & DeepSeek & Mar-24 \\
    & & WizardLM-2-8$\times$22B & WizardLM & Jun-24 \\
    & & Dolphin-Mixtral-8$\times$22B & Dolphin & Jan-24 \\
    \bottomrule
  \end{tabular}
  \label{tab:model_suite}
\end{table*}

\section{List of Models and Tasks}\label{ap:models_tasks}
In Table~\ref{tab:model_suite} and Table~\ref{tab:task}, we provide a detailed overview of all models and task intents included in FineServe.

\begin{table*}[htbp]
  \centering
  \caption{Task intent definitions used in our analysis.}
    \begin{tabular}{c|l}
    \toprule
    \textbf{Intent} & \textbf{Description} \\
    \midrule
    \textbf{Science} & Scientific reasoning, mathematics, engineering \\
    \textbf{Writing} & Essay writing, summarization, rewriting \\
    \textbf{Roleplaying} & Roleplay, fictional personas, interactive storytelling and adult \\
    \textbf{Entertainment} & Games, jokes, creative content, and casual entertainment \\
    \textbf{Social} & Conversation, personal interaction, and social queries \\
    \textbf{Finance} & Financial analysis, investment, accounting, and business planning \\
    \textbf{Health} & Medical, wellness, and health-related queries \\
    \textbf{Programme} & Code generation, debugging, and software-related tasks \\
    \textbf{Law} & Legal interpretation, compliance, and regulation-related queries  \\
    \textbf{Commerce} & E-commerce, product comparison, marketing, and sales-related queries\\
    \bottomrule
    \end{tabular}%
  \label{tab:task}%
\end{table*}%

\end{document}